\documentclass[a4paper, 10pt, conference]{ieeeconf}     
\usepackage{FG2026}

\FGfinalcopy 

\IEEEoverridecommandlockouts                         
\usepackage{times}
\usepackage{epsfig}
\usepackage{graphicx}
\usepackage{amsmath,amssymb,amsfonts,mathtools,bm}
\usepackage{booktabs,multirow}
\usepackage{siunitx}
\let\labelindent\relax
\usepackage{enumitem}
\usepackage[dvipsnames]{xcolor}
\usepackage{wrapfig}
\usepackage{adjustbox}
\usepackage[capitalize,noabbrev]{cleveref}
\usepackage{url}
\usepackage{nicefrac}
\usepackage{capt-of}
\usepackage{microtype}
\usepackage{xspace}
\usepackage{gensymb}
\usepackage{arydshln} 
\usepackage{multirow} 
\usepackage{makecell}

\usepackage{cite}

\usepackage{algorithm}
\usepackage{algpseudocode}

\newcommand{\SO}{\mathrm{SO}(3)}

\newcommand{\vect}[1]{\bm{#1}}
\newcommand{\mat}[1]{\bm{#1}}

\newcommand{\proj}{\pi}

\usepackage{pifont}
\newcommand{\cmark}{\ding{51}}
\newcommand{\xmark}{\ding{55}}

\newcommand{\eg}{e.g.,\xspace}
\newcommand{\ie}{i.e.,\xspace}

\title{\LARGE \bf
Multi-Camera Self-Calibration in Sports Motion Capture: \\Leveraging Human and Stick Poses
}

\author{\parbox{16cm}{\centering
    {\large Fan Yang$^1$, Changsoo Jung$^{1,2}$, Ryosuke Kawamura$^1$, and Hon Yung Wong$^1$}\\
    {\normalsize
    $^1$ Fujitsu Research\\
    $^2$ Colorado State University}}
}

\begin{document}

\maketitle
\thispagestyle{plain}

\begin{abstract}
Multi-camera systems are widely employed in sports to capture the 3D motion of athletes and equipment, yet calibrating their extrinsic parameters remains costly and labor-intensive. We introduce an efficient, tool-free method for multi-camera extrinsic calibration tailored to sports involving stick-like implements (\eg golf clubs, bats, hockey sticks). Our approach jointly exploits two complementary cues from synchronized multi-camera videos: (i) human body keypoints with unknown metric scale and (ii) a rigid stick-like implement of known length. We formulate a three-stage optimization pipeline that refines camera extrinsics, reconstructs human and stick trajectories, and resolves global scale via the stick-length constraint. Our method achieves accurate extrinsic calibration without dedicated calibration tools. To benchmark this task, we present the first dataset for multi-camera self-calibration in stick-based sports, consisting of synthetic sequences across four sports categories with 3 to 10 cameras. Comprehensive experiments demonstrate that our method delivers SOTA performance, achieving low rotation and translation errors. Our project page: \url{https://fandulu.github.io/sport_stick_multi_cam_calib/}.
\end{abstract}

\section{INTRODUCTION}

Multi-camera systems are essential for many sports analytics~\cite{wu2008multi,ren2010multi,luo2013feature, zhang2020multi,wang2024diffusion,yang2024enhancing,noorbhai2025conceptual,taylor2025biomechanical}, enabling precise 3D motion capture of athletes and handheld equipment (\eg baseball bats). This capability forms the foundation of 3D body movement recognition and offers valuable insights for performance evaluation and biomechanical analysis. A critical requirement for such applications is camera extrinsic calibration, which determines the relative positions and orientations of all cameras within a shared coordinate system. While intrinsic calibration is important, intrinsic parameters are typically fixed and require calibration only once, or can often be obtained directly from camera specifications. Conversely, extrinsic parameters often change whenever cameras are repositioned in a new environment, making recalibration both frequent and labor-intensive. Our method is specifically designed to tackle this challenge and support downstream sports activity analysis.

\begin{figure}[th!]
  \centering
  \includegraphics[width=\linewidth]{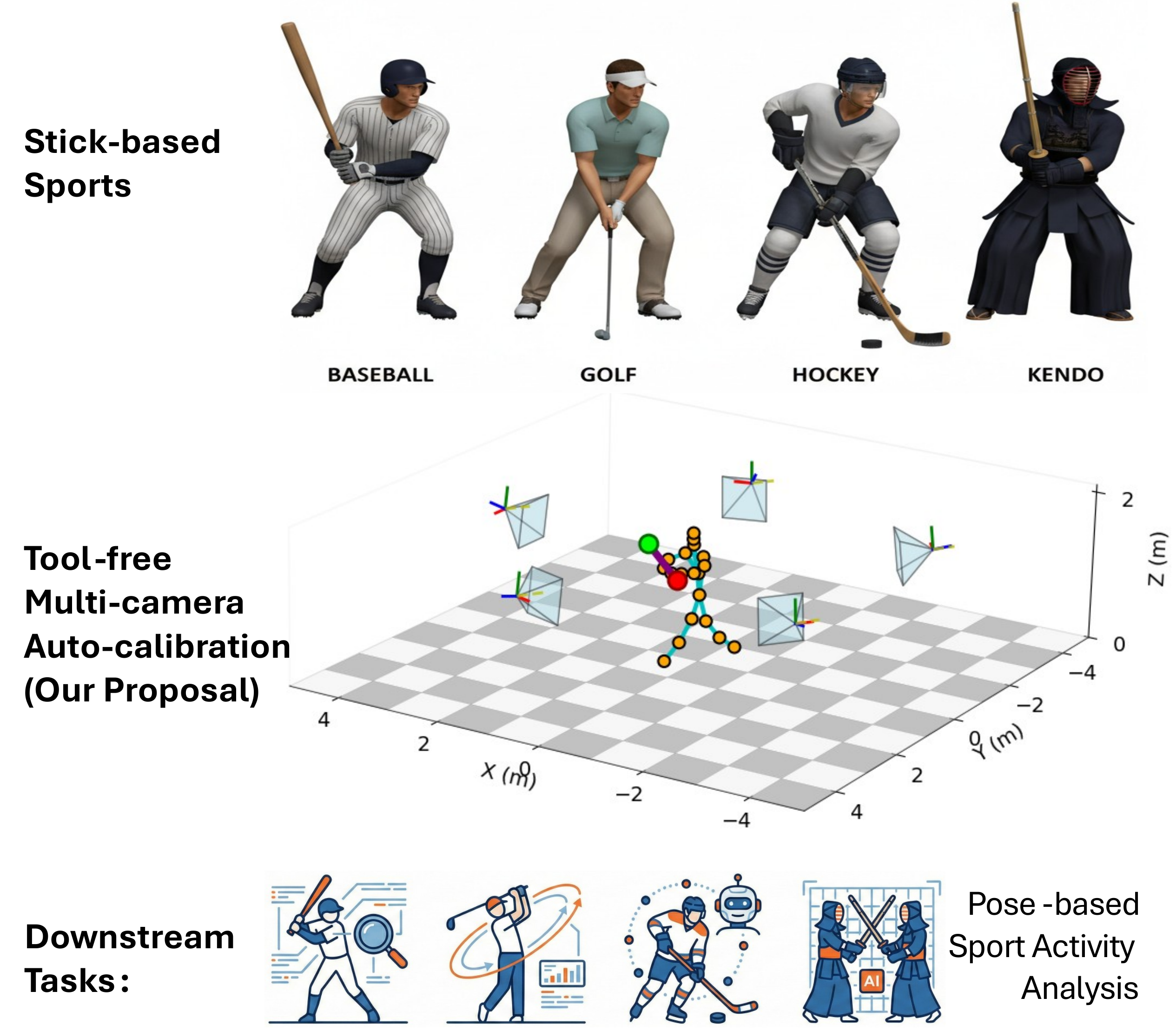}
  \caption{\textbf{Illustration of our proposal.} For sports involving stick-like implements (\eg golf, baseball, hockey, and kendo), we perform a novel tool-free multi-camera extrinsic calibration by leveraging both the stick and the human poses.}
  \label{fig:demo}
\end{figure}

Conventional calibration methods rely on dedicated physical tools, such as checkerboards or wand rigs~\cite{tabb2019multi, usenko2018double, xing2017new}. However, fabricating and transporting these specialized calibration tools to outdoor environments can be burdensome, particularly for sports capture scenarios involving frequent changes in camera position. Recent efforts have explored tool-free calibration using naturally occurring human motion~\cite{lee2022extrinsic,xu2021wide}, which eliminates the need for physical artifacts. However, these methods suffer from scale ambiguity since human body proportions vary across individuals and are rarely known a priori. While some approaches attempt to resolve scale using assumptions about average height~\cite{chauve2010robust} or rely on additional sensors such as LiDAR or IMUs~\cite{lee2021robust, yang2024yowo}, these solutions introduce significant cost, complexity, and synchronization requirements. Learning-based methods~\cite{allegro2025calib3r, keetha2025mapanything} also require domain-specific training and often lack guarantees of metric accuracy.

\begin{table*}[th!]
\centering
\caption{Comparison of multi-camera calibration methods. Our approach achieves real-scale extrinsic parameters without extra calibration tools, sensors, and learning.}
\setlength{\tabcolsep}{2.3pt}
\begin{tabular}{lcccc}
\toprule
\textbf{Method} & \textbf{Tool-free} & \textbf{Metric scale} & \textbf{RGB-only} & \textbf{Learning-free} \\
\midrule
Tool-based Methods (w/ Checkerboard / Wand)~\cite{gossard2024ewand,kim2018charuco,prible2024caliscope} & \xmark & \cmark & \cmark & \cmark \\
Tool-free Methods (w/ Extra Sensor)~\cite{desai2018skeleton, lee2021robust, yang2024yowo}                      & \cmark & \cmark & \xmark & \cmark \\
Tool-free Methods (w/ Learning-based SfM/Human Poses)~\cite{keetha2025mapanything,wang2025vggt, muller2025reconstructing, javerliat2025kineo}                      & \cmark & \xmark & \cmark & \xmark \\
Tool-free Methods (w/ Human Poses Only)~\cite{garau2020fast,lee2025spatiotemporal,lee2022extrinsic,pan2024global,patzold2022online,xu2021wide} & \cmark & \xmark & \cmark & \cmark \\
\textbf{Ours}                    & \textcolor{green}{\cmark} & \textcolor{green}{\cmark} & \textcolor{green}{\cmark} & \textcolor{green}{\cmark} \\
\bottomrule
\end{tabular}
\label{tab:related}
\end{table*}

We propose a tool-free calibration method that achieves metric accuracy by exploiting a common structure in many sports: a rigid, stick-like object of known length (\eg a golf club, hockey stick, or baseball bat). These objects provide a natural metric constraint and are often visible across multiple views. By combining the articulated motion of the athlete with the metrically constrained motion of the stick, we transform an under-constrained calibration problem into a well-posed one, without requiring any calibration hardware or auxiliary sensors.

Our approach jointly estimates camera extrinsics and reconstructs 3D trajectories of both the human and the stick using only 2D keypoints extracted from synchronized RGB videos. We formulate a three-stage optimization pipeline that reconstructs human and stick motion, refines camera parameters, and resolves global scale via the known stick length. To support evaluation, we introduce the first dataset for multi-camera calibration in stick-based sports, consisting of synthetic sequences with diverse motions, camera counts, and levels of noise.

Our contributions are threefold:
\begin{enumerate}%
\item \textbf{Benchmark for Scale-Aware Calibration:} We introduce the first benchmark for evaluating multi-camera extrinsic calibration using human and stick poses, encompassing diverse camera configurations, varied human and stick motions, and multiple noise levels.
\item \textbf{Tool-Free Multi-Camera Calibration for Sports:} We formulate the extrinsic calibration problem as a scale-aware bundle adjustment that incorporates metric constraints from any stick-like sports implement, enabling self-calibration without dedicated tools. This enables diverse downstream studies in multi-camera sports gesture analysis.
\item \textbf{Comprehensive Empirical Validation:} Across diverse configurations, our method consistently outperforms both traditional tool-based and recent tool-free baselines, achieving state-of-the-art (SOTA) rotation and translation accuracy.
\end{enumerate}


\section{Related Work}

\subsection{Calibration with Dedicated Tools} \label{sec:related_work_tools}
Multi-camera calibration systems using dedicated tools leverage the known geometric properties of those tools \cite{bolsee2020device, wu2022survey, gossard2024ewand,kanai2023robust,kedilioglu2025pricosa,kim2018charuco,pribanic2009comparison,prible2024caliscope,rameau2022mc,tabb2019multi,tripicchio2022multi}. Static objects provide precise and stable geometric scales, making tool-based approaches straightforward and reliable. For example, Jatesiktat \textit{et al.} utilize two LED wands to achieve extrinsic parameters from waving the wands \cite{jatesiktat2024multi}. Yuhai \textit{et al.} employ a T-wand and leverage the dynamic errors of the wand for accurate 3D reconstruction \cite{yuhai2024enhanced}. In addition, a double-sided checkerboard is applied to measure the target displacement error for solving multi-camera calibration \cite{marcon2017multicamera}. Xing \textit{et al.} employ special patterns on a checkerboard as CALTag \cite{atcheson2010caltag} to avoid symmetry ambiguities and achieve distinctive features for calibration \cite{xing2017new}.
While checkerboards, wands, and pattern rigs are widely used for camera calibration~\cite{jatesiktat2024multi, strauss2014calibrating, tabb2019multi, usenko2018double, xing2017new}, their construction and deployment impose practical constraints in outdoor sports scenarios.

\subsection{Marker-free Calibration from Human Motions} \label{sec:related_work_human_motions}
The human motion-based methods explore scenes in which people move around to establish correspondences across the views \cite{garau2020fast,lee2025spatiotemporal,lee2022extrinsic, patzold2022online,xu2021wide}. For example, Xu \textit{et al.} \cite{xu2021wide} demonstrate calibration using the motions of multiple people. They first employ human bounding boxes across cameras. Then, they mark each person as a keypoint, and the information of the keypoints is used for re-identifying the same person over the frames. The bounding-box correspondences are converted to point correspondences to solve camera poses~\cite{pan2024global}. Similarly, Lee \textit{et al.}~\cite{lee2022extrinsic} leverage movements of human body skeletons in room-scale calibration. The estimated 2D body joints are reconstructed as 3D body skeletons to compute extrinsic camera parameters. In this process, a trajectory of an oriented point set is collected from a freely moving person (\eg a point of the neck with a direction toward the spine), and it allows the method to predict the locations and poses of cameras. In addition, Kanai \textit{et al.} leverage self-supervised monocular depth and ego-motion to learn extrinsic parameters with other scene properties \cite{kanai2023robust}.

However, methods that utilize human pose without metric constraints determine only up to a global similarity, leaving the scale undetermined. Although Guan \textit{et al.}~\cite{chauve2010robust} assume an average height for all observed pedestrians to resolve scale, the limitation is obvious: human heights vary significantly, making this an inaccurate constraint. A 10\% error in assumed height could translate to a more than 10\% error in the global scale. Our method leverages the fixed-length stick inherent in sports settings to substantially reduce scale ambiguity.

\subsection{Calibration with Extra Sensors Beyond RGB Cameras} \label{sec:related_work_egocentric}
A prevalent strategy for resolving the inherent scale ambiguity in multi-camera systems is the integration of auxiliary sensors that provide direct metric measurements of the environment. Introducing sensors such as LiDAR or IMUs can anchor the extrinsic calibration in a real-world coordinate frame, thereby recovering the absolute metric scale through fusion~\cite{desai2018skeleton, lee2021robust, yang2024yowo}. 
However, this sensor fusion approach introduces significant practical challenges. It requires additional, often costly hardware and imposes a stringent need for precise spatiotemporal synchronization across heterogeneous data streams. Even minor timing misalignments can lead to substantial calibration errors, especially in dynamic scenes~\cite{lee2021robust}. These drawbacks, such as high cost, complexity, and strict synchronization demands, render multi-modal systems impractical for many applications, particularly sports analytics, which prioritizes rapid, scalable, and non-intrusive setups. Our approach obviates these issues by relying solely on RGB data, which makes our solution a more feasible and scalable alternative for both indoor and outdoor sports capture.

\subsection{Calibration with Learning-based Structure From Motion} \label{sec:related_work_learning_based}
While some prior work relies on geometric correspondences, learning-based Structure-from-Motion (SfM) approaches have been extensively explored. These methods are typically trained on large-scale datasets and perform end-to-end estimation of multi-camera extrinsic parameters. For example, Allegro et al.~\cite{allegro2025calib3r} employ a 3D foundation model to perform SfM from RGB data combined with robot pose information. Other recent studies~\cite{keetha2025mapanything,wang2025vggt} leverage transformer-based architectures to infer extrinsic parameters directly from images or videos. Although these approaches demonstrate strong performance, they often lack guarantees of metric scale, frequently require domain-specific training, and may degrade under novel or challenging conditions. In contrast, our method is training-free and exhibits stronger generalization. Recent works~\cite{muller2025reconstructing,javerliat2025kineo} utilize human pose information to refine learning-based calibration results; however, they do not focus on specific sports scenarios nor exploit sport-specific equipment as a geometric reference.

\subsection{Multi-Camera Calibration in Sports} \label{sec:related_work_calib_sports}
In some sports, field markers and lines are leveraged for multi-camera calibration. For example, in soccer, field lines and segment correspondences are commonly used to optimize camera poses and focal lengths~\cite{chen2019sports, theiner2023tvcalib}. Similarly, Citraro \textit{et al.}~\cite{citraro2020real} utilize court lines in basketball to compute 3D correspondences by detecting 2D line intersections, which are then used to estimate camera poses. However, these methods primarily depend on sport-specific field markings or calibration features embedded in the environment. Such assumptions limit their applicability in arbitrary scenes that lack metric ground markers, such as in golf or some indoor sports. Our method overcomes these limitations and can be applied to a broader range of multi-camera sports capture scenarios.

\section{Method}

\begin{figure}[th!]
  \centering
  \includegraphics[width=\linewidth]{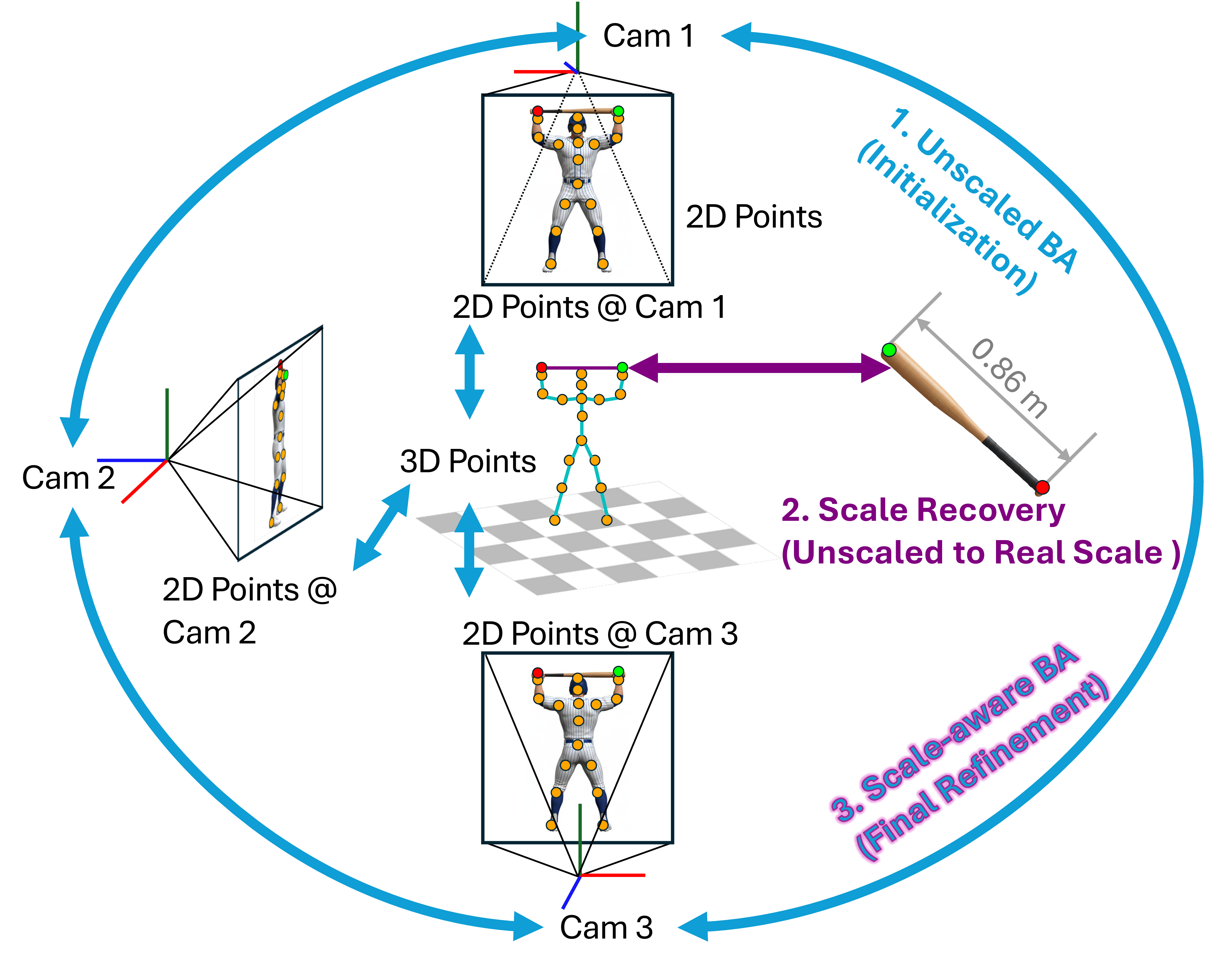}
  \caption{\textbf{Three-stage optimization.} (1) An initial, unscaled 3D pose is reconstructed from multi-view 2D keypoints via Bundle Adjustment (BA). (2) The real-world scale is recovered using a known measurement, such as the length of the baseball bat (0.86 m). (3) A final, scale-aware BA is performed to refine the metric 3D reconstruction.}
  \label{fig:method}
\end{figure}

Our method solves for camera extrinsics and 3D human-stick trajectories by jointly optimizing over multi-view observations. The core idea is to first establish a geometrically consistent but scale-ambiguous reconstruction using all available keypoints (human and stick), and then resolve the metric scale using the known length of the stick. This is achieved via a three-stage pipeline: unscaled bundle adjustment, scale recovery, and a final scale-aware bundle adjustment with length constraints, as shown in Fig.~\ref{fig:method}. The complete pipeline is outlined in Algorithm~\ref{alg:method}. We assume synchronized cameras with known intrinsics, overlapping cross-camera views, and reliable 2D detections for a subset of frames. The length of the stick-like implement is determined based on standard dimensions commonly used in the corresponding sports.

\subsection{Preliminaries and Notation}

We consider a setup of $C$ synchronized cameras observing a person with a stick over $F$ frames. The intrinsic matrix $\mat{K}_i$ of each camera $i$ is assumed to be known. Our goal is to estimate the extrinsic parameters: rotation $\mat{R}_i \in \SO$ and translation $\vect{t}_i \in \mathbb{R}^3$, for each camera.

The 3D structure consists of $J$ human keypoints and two stick endpoints. At frame $f$, their 3D positions in the world coordinate system are denoted by $\{\vect{Y}_{f,j} \in \mathbb{R}^3\}_{j=1}^J$ for the human and $\{\vect{X}_{f,e} \in \mathbb{R}^3\}_{e=0}^1$ for the stick. The corresponding 2D observations in camera $i$ are $\{\vect{v}_{i,f,j} \in \mathbb{R}^2\}$ and $\{\vect{u}_{i,f,e} \in \mathbb{R}^2\}$.

\paragraph{Visibility Masks}
In practice, not all keypoints are visible in every camera view due to occlusions or detection failures. We define binary visibility masks $\mat{M}_{i,f,j} \in \{0,1\}$ and $\mat{N}_{i,f,e} \in \{0,1\}$ to indicate whether human keypoint $j$ and stick endpoint $e$ are visible in camera $i$ at frame $f$. These masks are used to selectively include valid observations in triangulation, reprojection loss, and Jacobian construction.

\paragraph{Projection Model}
The projection of a 3D point $\vect{P} \in \mathbb{R}^3$ onto the $i$-th camera's image plane is given by:
\begin{equation} \footnotesize 
\label{eq:projection}
\tilde{\vect{p}} = \mat{K}_i (\mat{R}_i \vect{P} + \vect{t}_i), \quad
\proj(\mat{K}_i, \mat{R}_i, \vect{t}_i, \vect{P}) = \frac{1}{\tilde{p}_z} \begin{bmatrix} \tilde{p}_x \\ \tilde{p}_y \end{bmatrix}
\end{equation}
To fix the global coordinate system, we set the first camera as the reference: $\mat{R}_1 = \mat{I}, \vect{t}_1 = \vect{0}$.

\subsection{Initialization}

\begin{figure*}[th!]
  \centering
  \includegraphics[width=\linewidth]{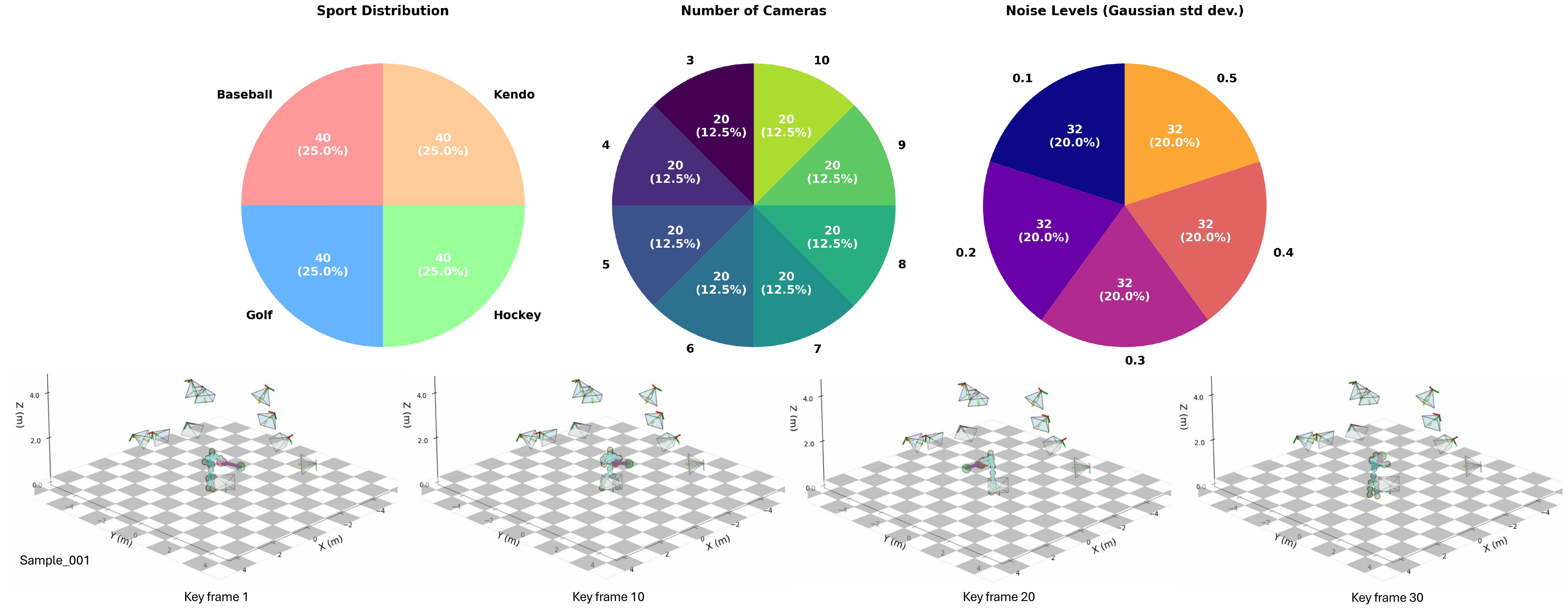}
  \caption{\textbf{Our Sports-Stick-Syn dataset.} \textbf{Top:} Statistics of our dataset, including sports categories, number of cameras, and noise levels. \textbf{Bottom:} An example visualization from the dataset.}
  \label{fig:data}
\end{figure*}

\begin{algorithm}[h!]
\footnotesize
\caption{Multi-Camera Calibration with Metric Stick} \label{alg:method}
\begin{algorithmic}[1]
\Require Intrinsics $\{\mat{K}_i\}$, 2D human points $\{\vect{v}_{i,f,j}\}$, 2D stick endpoints $\{\vect{u}_{i,f,e}\}$, known stick length $L$.
\State \textbf{Initialize:}
\State Fix gauge: $\mat{R}_1 \leftarrow \mat{I}, \vect{t}_1 \leftarrow \vect{0}$.
\State Compute pairwise essential matrices $\{\mat{E}_{ij}\}$ for overlapping views.
\State Initialize $\{\mat{R}_i, \vect{t}_i\}$ by propagating poses along a maximum spanning tree rooted at Camera 1.
\State Triangulate initial $\{\vect{Y}_{f,j}\}$ and $\{\vect{X}_{f,e}\}$ using all visible views.
\State \textbf{Stage 1: Unscaled Joint BA}
\State Minimize $\mathcal{L}_{\text{reproj}}$ (Eq.~\eqref{eq:Lreproj}) over $\{\mat{R}_i,\vect{t}_i\},\{\vect{Y}_{f,j}\},\{\vect{X}_{f,e}\}$.
\State \textbf{Stage 2: Scale Recovery}
\State Compute $\bar{L}=\frac{1}{F}\sum_f \|\hat{\vect{X}}_{f,1}-\hat{\vect{X}}_{f,0}\|_2$ and $s=L/\bar{L}$.
\State Rescale: $\vect{t}_i \leftarrow s\vect{t}_i,\ \vect{Y}_{f,j}\leftarrow s\vect{Y}_{f,j},\ \vect{X}_{f,e}\leftarrow s\vect{X}_{f,e}$.
\State \textbf{Stage 3: Scale-aware BA}
\State Minimize Eq.~\eqref{eq:ba3_endpoints} with stick-length and temporal-smoothness constraints.
\State \Return Calibrated extrinsics $\{(\mat{R}_i,\vect{t}_i)\}_{i=1}^C$ and 3D trajectories.
\end{algorithmic}
\end{algorithm}

\paragraph{Camera Pose Estimation}

We initialize camera extrinsics by traversing the inter-camera overlap graph. For each overlapping pair $(i,j)$, we estimate the essential matrix $\mat{E}_{ij}$ with RANSAC using all detected keypoints (human and stick), then decompose it into $(\mat{R}_{ij},\vect{t}_{ij})$ and select the valid solution via a cheirality check. We construct a maximum spanning tree weighted by inlier counts and propagate pairwise poses along the tree to obtain $(\mat{R}_i,\vect{t}_i)$ relative to Camera~1. Translations are recovered up to an unknown global scale.

\paragraph{3D Structure Triangulation}
With the initial camera poses, we triangulate the initial 3D position of every human keypoint $\vect{Y}_{f,j}$ and stick endpoint $\vect{X}_{f,e}$ for each frame $f$. This is achieved using Direct Linear Transformation (DLT) with the 2D observations from all visible cameras (\ie those for which $\mat{M}_{i,f,j} = 1$ or $\mat{N}_{i,f,e} = 1$).

\subsection{Stage 1: Unscaled Joint Bundle Adjustment}

The initial estimates are refined in a joint bundle adjustment step. This stage optimizes camera extrinsics $(\{\mat{R}_i, \vect{t}_i\}_{i=2}^C)$ and the 3D trajectories of all keypoints $(\{\vect{Y}_{f,j}\}, \{\vect{X}_{f,e}\})$ to minimize a robust, visibility-masked reprojection loss:
\begin{equation}
\footnotesize
\label{eq:Lreproj}
\begin{aligned}[t]
\mathcal{L}_{\text{reproj}}
(\{\mat{R}_i,\vect{t}_i\},\{\vect{Y}_{f,j}\},\{\vect{X}_{f,e}\})
&=
\sum_{i=1}^{C}\sum_{f=1}^{F}
\Bigg[
\sum_{j=1}^{J}
\mat{M}_{i,f,j}\,
\rho\!\left(\left\|\vect{r}_{i,f,j}^{H}\right\|_2^2\right)
\\
&\quad\;
+
\sum_{e\in\{0,1\}}
\mat{N}_{i,f,e}\,
\rho\!\left(\left\|\vect{r}_{i,f,e}^{S}\right\|_2^2\right)
\Bigg]
\end{aligned}
\end{equation}
where the residuals are
\begin{equation} \footnotesize
\label{eq:residuals_stage1}
\begin{split}
\vect{r}_{i,f,j}^H &= \vect{v}_{i,f,j} - \proj(\mat{K}_i, \mat{R}_i, \vect{t}_i, \vect{Y}_{f,j}), \nonumber \\
\vect{r}_{i,f,e}^S &= \vect{u}_{i,f,e} - \proj(\mat{K}_i, \mat{R}_i, \vect{t}_i, \vect{X}_{f,e}).
\end{split}
\end{equation}
and $\rho(\cdot)$ is a robust loss function (e.g., Cauchy) to mitigate outliers from 2D detection errors.
In this stage, we aim to minimize $\mathcal{L}_{\text{reproj}}$.

\subsection{Stage 2: Closed-Form Scale Recovery}

The global scale of the reconstruction is resolved using the known physical length, $L$, of the stick. We compute the average reconstructed length $\bar{L}$ from the optimized 3D stick endpoints:
\begin{equation} \footnotesize
\bar{L} = \frac{1}{F} \sum_{f=1}^{F} \left\|\hat{\vect{X}}_{f,1} - \hat{\vect{X}}_{f,0}\right\|_2,
\qquad
s = \frac{L}{\bar{L}}.
\end{equation}
We then rescale all translational components:
\begin{equation} \footnotesize
\vect{t}_i \leftarrow s\vect{t}_i,\qquad
\vect{Y}_{f,j} \leftarrow s\vect{Y}_{f,j},\qquad
\vect{X}_{f,e} \leftarrow s\vect{X}_{f,e}.
\end{equation}

\subsection{Stage 3: Scale-Aware Bundle Adjustment}

This stage refines camera extrinsics and 3D trajectories in metric space.
We reuse the reprojection objective $\mathcal{L}_{\text{reproj}}$ from Eq.~\eqref{eq:Lreproj} and add (i) a stick-length constraint and (ii) temporal smoothness.

\paragraph{Stick Length Constraint}
We enforce rigidity of the stick by penalizing deviation from its known length $L$:
\begin{equation} \footnotesize
\label{eq:Llength_stick}
\mathcal{L}_{\text{length}}^{S} = \sum_{f=1}^{F} \left( \left\|\vect{X}_{f,1} - \vect{X}_{f,0}\right\|_2 - L \right)^2.
\end{equation}

\paragraph{Temporal Smoothness}
To encourage smooth motion, we penalize second-order differences (acceleration) across frames:
\begin{equation} \footnotesize
\label{eq:Lsmooth}
\begin{split}
\mathcal{L}_{\text{smooth}} = 
&\sum_{f=2}^{F-1}\sum_{j=1}^{J}
\left\|\vect{Y}_{f-1,j} - 2\vect{Y}_{f,j} + \vect{Y}_{f+1,j}\right\|_2^2 \\
&+ \sum_{f=2}^{F-1}\sum_{e\in\{0,1\}}
\left\|\vect{X}_{f-1,e} - 2\vect{X}_{f,e} + \vect{X}_{f+1,e}\right\|_2^2.
\end{split}
\end{equation}

\paragraph{Final Optimization Objective}
We jointly optimize camera extrinsics $\{(\mat{R}_i, \vect{t}_i)\}$, human keypoints $\{\vect{Y}_{f,j}\}$, and stick endpoints $\{\vect{X}_{f,e}\}$ by minimizing:
\begin{equation} \footnotesize
\label{eq:ba3_endpoints}
\min_{\substack{\{\mat{R}_i,\vect{t}_i\}_{i=2}^C\\ \{\vect{Y}_{f,j}\},\{\vect{X}_{f,e}\}}}
\mathcal{L}_{\text{reproj}}
+ \lambda_{\text{length}}\,\mathcal{L}_{\text{length}}^{S}
+ \lambda_{\text{smooth}}\,\mathcal{L}_{\text{smooth}}, 
\end{equation}
where $\lambda_{\text{length}}$ and $\lambda_{\text{smooth}}$ are heuristically set as 0.4 and 0.2, respectively.

\paragraph{Sparse Jacobian Construction}
We minimize a sum of (robust) squared residuals that includes reprojection, stick-length, and temporal smoothness terms. The Jacobian matrix $\mat{J}$ is sparse because each residual depends only on a small subset of variables.

\begin{itemize}
\item \textbf{Reprojection residuals (visible points only).}
In practice, $\rho$ is handled via IRLS, yielding per-residual weights; the Jacobians below are for the underlying residuals. For a human keypoint $j$ at frame $f$ in camera $i$:
\begin{equation}\footnotesize
\vect{r}_{i,f,j}^{H}
=
\vect{v}_{i,f,j}
-
\proj(\mat{K}_i,\mat{R}_i,\vect{t}_i,\vect{Y}_{f,j}),
\qquad \text{if } \mat{M}_{i,f,j}=1.
\end{equation}
For a stick endpoint $e\in\{0,1\}$ at frame $f$ in camera $i$:
\begin{equation}\footnotesize
\vect{r}_{i,f,e}^{S}
=
\vect{u}_{i,f,e}
-
\proj(\mat{K}_i,\mat{R}_i,\vect{t}_i,\vect{X}_{f,e}),
\qquad \text{if } \mat{N}_{i,f,e}=1.
\end{equation}

Let $\mat{J}_{\proj}(\vect{P}) \in \mathbb{R}^{2\times 3}$ denote the Jacobian of the projection function w.r.t.\ the 3D point $\vect{P}$:
\begin{equation}\footnotesize
\mat{J}_{\proj}(\vect{P}) \;=\; \frac{\partial}{\partial \vect{P}}
\proj(\mat{K}_i,\mat{R}_i,\vect{t}_i,\vect{P}).
\end{equation}
Then the Jacobians w.r.t.\ the 3D point variables are
\begin{equation}\footnotesize
\frac{\partial \vect{r}_{i,f,j}^{H}}{\partial \vect{Y}_{f,j}}
=
-\mat{J}_{\proj}(\vect{Y}_{f,j}),
\qquad
\frac{\partial \vect{r}_{i,f,e}^{S}}{\partial \vect{X}_{f,e}}
=
-\mat{J}_{\proj}(\vect{X}_{f,e}),
\end{equation}
with all other derivatives w.r.t.\ unrelated points equal to zero.

Each reprojection residual also depends on the extrinsics $(\mat{R}_i,\vect{t}_i)$ of camera $i$. Using a minimal perturbation $\delta \boldsymbol{\xi}_i \in \mathbb{R}^6$ on $\mathrm{SE}(3)$ (e.g., $\mat{R}_i \leftarrow \exp([\delta\boldsymbol{\omega}_i]_\times)\mat{R}_i$, $\vect{t}_i \leftarrow \vect{t}_i+\delta\vect{t}_i$), we include the corresponding blocks
\begin{equation}\footnotesize
\frac{\partial \vect{r}_{i,f,\cdot}}{\partial \delta \boldsymbol{\xi}_i}
\;=\;
-\frac{\partial}{\partial \delta \boldsymbol{\xi}_i}
\proj(\mat{K}_i,\mat{R}_i,\vect{t}_i,\vect{P}),
\end{equation}
where $\vect{P}$ is $\vect{Y}_{f,j}$ or $\vect{X}_{f,e}$. These blocks are non-zero only for residuals from camera $i$.

\item \textbf{Stick-length residuals.}
We enforce the known stick length $L$ using the scalar residual
\begin{equation}\footnotesize
r_f^{\text{len}}
=
\left\|\vect{X}_{f,1}-\vect{X}_{f,0}\right\|_2 - L.
\end{equation}
Let $\vect{d}_f = \vect{X}_{f,1}-\vect{X}_{f,0}$ and $\ell_f=\|\vect{d}_f\|_2$. The Jacobian blocks are
\begin{equation}\footnotesize
\frac{\partial r_f^{\text{len}}}{\partial \vect{X}_{f,1}}
=
\frac{\vect{d}_f}{\ell_f},
\qquad
\frac{\partial r_f^{\text{len}}}{\partial \vect{X}_{f,0}}
=
-\frac{\vect{d}_f}{\ell_f},
\end{equation}
and derivatives w.r.t.\ all other variables are zero. We use $\ell_f \leftarrow \max(\ell_f,\epsilon)$ with $\epsilon \approx 10^{-6}$.

\item \textbf{Temporal smoothness residuals.}
For each human keypoint $j$:
\begin{equation}\footnotesize
\vect{r}_{f,j}^{\text{sm},H}
=
\vect{Y}_{f-1,j} - 2\vect{Y}_{f,j} + \vect{Y}_{f+1,j},
\qquad f=2,\dots,F-1.
\end{equation}
For each stick endpoint $e\in\{0,1\}$:
\begin{equation}\footnotesize
\vect{r}_{f,e}^{\text{sm},S}
=
\vect{X}_{f-1,e} - 2\vect{X}_{f,e} + \vect{X}_{f+1,e},
\qquad f=2,\dots,F-1.
\end{equation}
The Jacobian blocks for the human smoothness residual are
\begin{equation}\footnotesize
\frac{\partial \vect{r}_{f,j}^{\text{sm},H}}{\partial \vect{Y}_{f-1,j}}=\mat{I}_3,\quad
\frac{\partial \vect{r}_{f,j}^{\text{sm},H}}{\partial \vect{Y}_{f,j}}=-2\mat{I}_3,\quad
\frac{\partial \vect{r}_{f,j}^{\text{sm},H}}{\partial \vect{Y}_{f+1,j}}=\mat{I}_3,
\end{equation}
and similarly for the stick smoothness residual:
\begin{equation}\footnotesize
\frac{\partial \vect{r}_{f,e}^{\text{sm},S}}{\partial \vect{X}_{f-1,e}}=\mat{I}_3,\quad
\frac{\partial \vect{r}_{f,e}^{\text{sm},S}}{\partial \vect{X}_{f,e}}=-2\mat{I}_3,\quad
\frac{\partial \vect{r}_{f,e}^{\text{sm},S}}{\partial \vect{X}_{f+1,e}}=\mat{I}_3.
\end{equation}

\end{itemize}

Each reprojection residual contributes non-zero Jacobian blocks only to the parameters of (i) its camera $i$ and (ii) its associated 3D point at frame $f$. Each stick-length residual touches only the two stick endpoints $(\vect{X}_{f,0}, \vect{X}_{f,1})$ at that frame. Each temporal smoothness residual touches only three consecutive frames $(f-1,f,f+1)$ of the same keypoint or endpoint. Therefore, $\mat{J}$ is highly sparse, enabling efficient optimization.

\section{Experiments}
\label{sec:exp}

To validate our proposed method, we conduct a series of comprehensive experiments. Our evaluation is designed to answer the following key questions: (1) How accurate is our method in recovering metric camera extrinsics compared to existing approaches? (2) What is the contribution of each component in our proposed framework? (3) How robust is our method to challenging conditions, such as a varying number of cameras and noisy 2D detections? We first introduce our newly created synthetic benchmark and a real-world dataset, followed by detailed quantitative and qualitative analyses.

\subsection{Datasets and Setup}

Since no public dataset exists for this task, we created a novel synthetic dataset called \textbf{Sports-Stick-Syn} (see Fig.~\ref{fig:data}). We developed a novel simulator based on principles of human biomechanics to generate realistic scenes. The simulator includes four distinct sports: golf, baseball, hockey, and kendo. For each sport, we use a stick of standardized length as defined by the official regulations of the respective sports league. We attach this stick to a virtual character’s hands. The character moves uniformly across the capture space while performing a variety of motion patterns with the stick. For each motion sequence, we simulate multiple camera configurations by varying the number of cameras from 3 to 10, arranged in both semi-spherical and random layouts. This setup reflects a diverse range of real-world capture scenarios in sports arenas.

The synthetic nature of the dataset allows us to generate perfect ground-truth data. This includes all camera intrinsics and extrinsics, 3D human joint positions, and 3D stick endpoint trajectories for every frame. We then project the 3D points onto each camera’s image plane to obtain 2D keypoint detections. To simulate real-world noise in pose estimation~\cite{mmpose2020, yolo11_ultralytics}, we add varying levels of Gaussian noise to the 2D keypoint coordinates. To compare with traditional methods, we also created an ArUco cube~\cite{kim2018charuco} and placed it in the scene center to enable traditional extrinsic calibration~\cite{bolsee2020device, kedilioglu2025pricosa,kim2018charuco,rameau2022mc,tabb2019multi}.

\subsection{Evaluation Metrics}

To evaluate the accuracy of the estimated camera extrinsics, we perform a quantitative comparison. We align the reconstructed camera centers and orientations with the ground-truth values. For this alignment, we use the Kabsch-Umeyama algorithm~\cite{umeyama2002least} without scale, since scale is the parameter we aim to estimate. Let $\{\hat{\vect{t}_i}\}_{i=1}^C$ be the estimated camera centers and $\{\vect{t}_i^{\text{gt}}\}_{i=1}^C$ be the ground-truth camera centers. The goal is to find a rigid transformation $(\mat{R}_{\text{align}}, \vect{t}_{\text{align}})$ such that:
\begin{equation} \footnotesize
\min_{\mat{R}_{\text{align}}, \vect{t}_{\text{align}}} \sum_{i=1}^{C} \left\|  \mat{R}_{\text{align}} \cdot \hat{\vect{t}_i} + \vect{t}_{\text{align}} - \vect{t}_i^{\text{gt}} \right\|^2
\end{equation}
The Kabsch-Umeyama algorithm solves this in closed form. After alignment, we report the following metrics:
\begin{itemize}
    \item \textbf{Rotation Error}: Angular deviation between aligned and ground-truth rotations, computed as:
    \begin{equation} \footnotesize
    \theta_i = \arccos\left( \frac{\text{Tr}\left((\mat{R}_i^{\text{gt}})^\top (\mat{R}_{\text{align}} \hat{\mat{R}}_i)\right) - 1}{2} \right)
    \end{equation}
    \item \textbf{Translation Error}: Euclidean distance between aligned and ground-truth camera centers:
    \begin{equation} \footnotesize
    \delta_i = \left\|  \mat{R}_{\text{align}} \cdot \hat{\vect{t}_i} + \vect{t}_{\text{align}} - \vect{t}_i^{\text{gt}} \right\|
    \end{equation}
\end{itemize}
This alignment ensures that evaluation is invariant to global similarity transformations and isolates the accuracy of the estimated extrinsics.

\subsection{Comparison with existing SOTA approaches}
We select representative open-source SOTA approaches for comparison, including methods that use dedicated calibration tools and those that do not.
In the following evaluation, the best and second-best results are in \textbf{bold} and \underline{underlined}, respectively.

As reported in Tab.~\ref{tab:compare_SOTA}, our proposed method achieves SOTA performance, significantly outperforming existing approaches across all metrics. 
First, compared to the dedicated tool-based baseline ArucoCalib~\cite{tabb2019multi}, our method reduces average rotation and translation errors by approximately 77\% and 86\%, respectively. While ArucoCalib provides a reliable baseline, its accuracy is constrained by the static placement of the marker, which limits overlapping views and feature distribution.
Second, regarding learning-based and human-centric approaches, we observe distinct limitations in current SOTA methods. Methods relying solely on human poses (CalibPerson~\cite{lee2022extrinsic}, ExtraCalib~\cite{patzold2022online}) face inherent scale ambiguities. Even with ground-truth height priors ($h_{human}^{gt}$), the variance in effective human height during motion prevents precise translation recovery. Hybrid methods like HSfM~\cite{muller2025reconstructing} and Kineo~\cite{javerliat2025kineo} improve structure recovery, but still suffer from significant translation drift due to the lack of rigid geometric constraints.

\begin{table}[h!]
\centering
\caption{Comparison with existing SOTA approaches on our benchmark.}
\label{tab:compare_SOTA}
\setlength{\tabcolsep}{.3pt}
\begin{tabular}{lcc}
\toprule
\textbf{Approach} & \textbf{Avg Rot. Err. ($^\circ$)}$\downarrow$ & \textbf{Avg Trans. Err. (m)}$\downarrow$ \\
\midrule
 \multicolumn{3}{l}{\textcolor{gray}{w/ dedicated calibration tools}} \\
ArucoCalib~\cite{tabb2019multi} (arXiv 2019)  & 0.087   & \underline{0.007}  \\ 
\hdashline
\multicolumn{3}{l}{\textcolor{gray}{ w/ human poses only}} \\
CalibPerson~\cite{lee2022extrinsic}  (RA-L 2022) \\
~~w/ $h_{human}^{gt}$    &  0.268  & 0.072   \\
ExtraCalib~\cite{patzold2022online} (GCPR 2022) \\
~~w/ $h_{human}^{gt}$    &  0.239  & 0.044   \\
\hdashline
\multicolumn{3}{l}{\textcolor{gray}{w/ learning-based SfM + human-pose refinement}} \\
HSfM~\cite{muller2025reconstructing} (CVPR 2025) &  0.193  & 0.117 \\
Kineo~\cite{javerliat2025kineo} (arXiv 2025) &  \underline{0.075}  & 0.098\\
\hdashline
\textbf{Ours} & \textbf{0.020} & \textbf{0.001} \\
\bottomrule
\end{tabular}
\end{table}

\begin{figure}[th!]
  \centering
  \includegraphics[width=\linewidth]{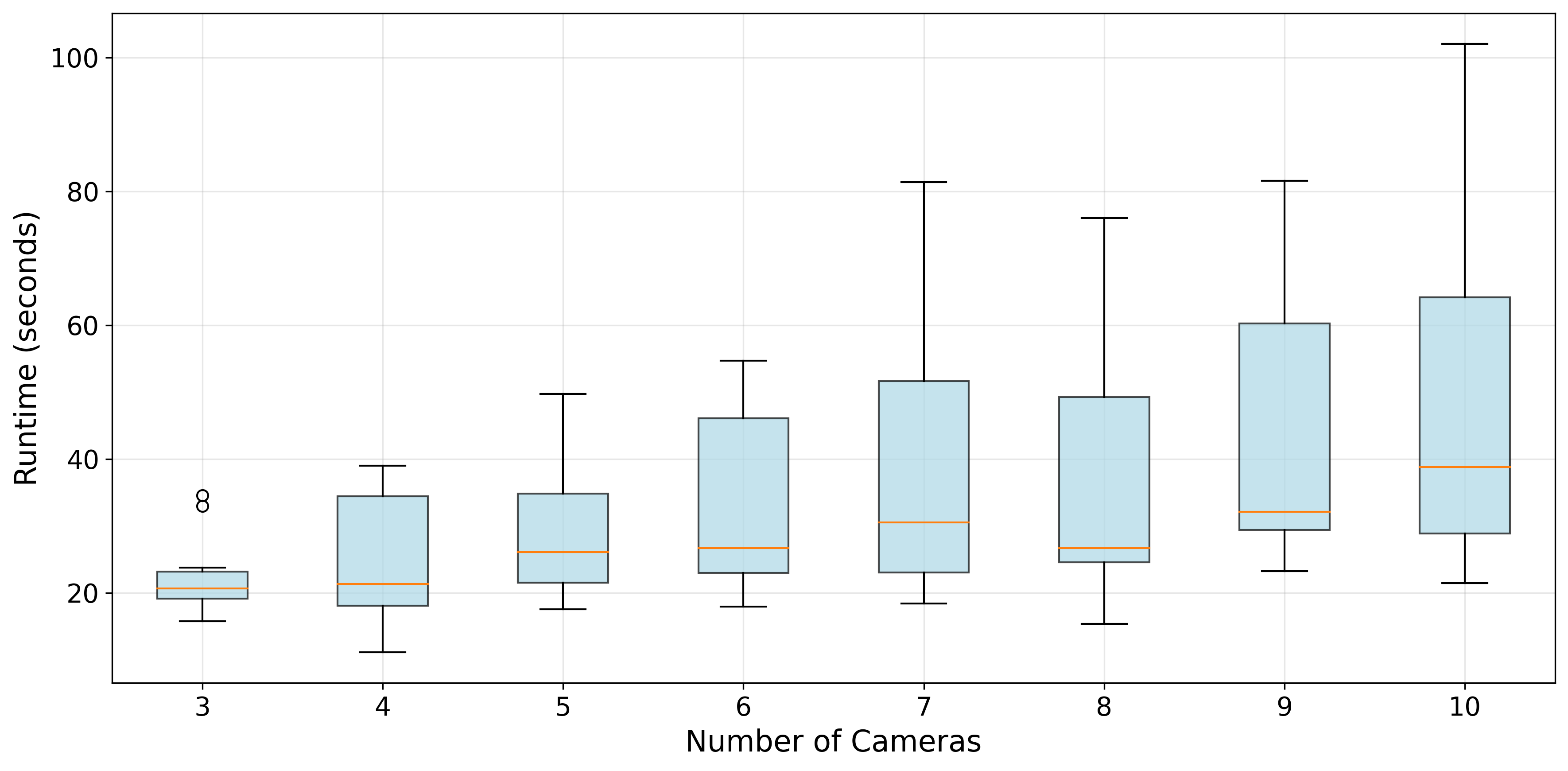}
  \vspace{-0.5cm}
  \caption{\textbf{Runtime distribution for our method.}  Each box plot shows the median (orange line), interquartile range, and outliers. }
  \label{fig:runtime_boxplot}
\end{figure}

\begin{table}[h!]
\centering
\setlength{\tabcolsep}{1pt}
\caption{Memory Usage Summary (MB)}
\vspace{-0.2cm}
\label{tab:memory}
\begin{tabular}{cccc}
\toprule
\textbf{Cams}  & \textbf{Initial Mem.} & \textbf{Peak Mem. (typical)} & \textbf{Peak Mem. (worst)} \\
\midrule
3  & 0.08 & 0.2 & 1.5 \\
4  & 0.09 & 0.2 & 2.1 \\
5  & 0.11 & 0.2 & 2.6 \\
6  & 0.13 & 0.3 & 3.2 \\
7  & 0.14 & 0.3 & 3.8 \\
8  & 0.16 & 0.4 & 4.4 \\
9  & 0.18 & 0.4 & 5.0 \\
10 & 0.19 & 0.5 & 5.6 \\
\bottomrule
\end{tabular}
\end{table}

\begin{table*}[h!]
  \centering
  \caption{Ablation Study: Input Variants. We report the average rotation and translation errors, along with their variances.}
  \label{tab:input_ablation}
  \setlength{\tabcolsep}{3.8pt}
  \begin{tabular}{lcccc}
    \toprule
    \textbf{Inputs Settings} & \textbf{Avg Rot. Err. ($^\circ$)}$\downarrow$ & \textbf{Avg Trans. Err. (m)}$\downarrow$ & \textbf{Avg Rot. Var.}$\downarrow$ & \textbf{Avg Trans. Var.}$\downarrow$ \\ 
    \midrule
    Human (w/ real human height $h_{human}^{gt}$) & 0.217 & 0.031 & 0.089 & 0.014 \\
    Human (w/ guessed height $h_{human}^{gt}\pm 0.1 m$) & 0.252 & 0.043 & 0.102 & 0.022 \\
    Stick (given scale) & \underline{0.023} & \textbf{0.001} & \underline{0.018} & \textbf{0.001} \\
    \textbf{Ours (Full)} human (no scale) + stick (given scale) & \textbf{0.020} & \textbf{0.001} & \textbf{0.014} & \textbf{0.001} \\
    \bottomrule
  \end{tabular}
\end{table*}

\begin{figure*}[th!]
  \centering
  \includegraphics[width=\linewidth]{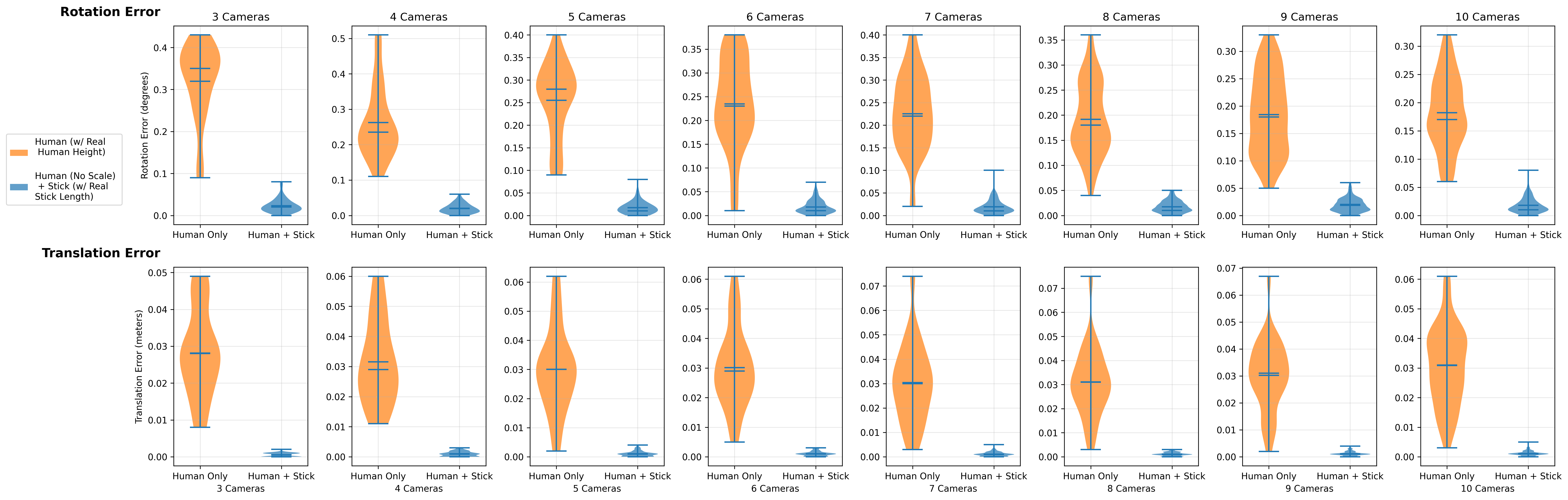}
  \vspace{-0.6cm}
  \caption{\textbf{Per-camera Error Distribution.} The human + stick approach yields lower errors and variance in both rotation and translation. }
  \label{fig:camera_error_distributions}
\end{figure*}

\begin{figure*}[th!]
  \centering
  \includegraphics[width=0.85\linewidth]{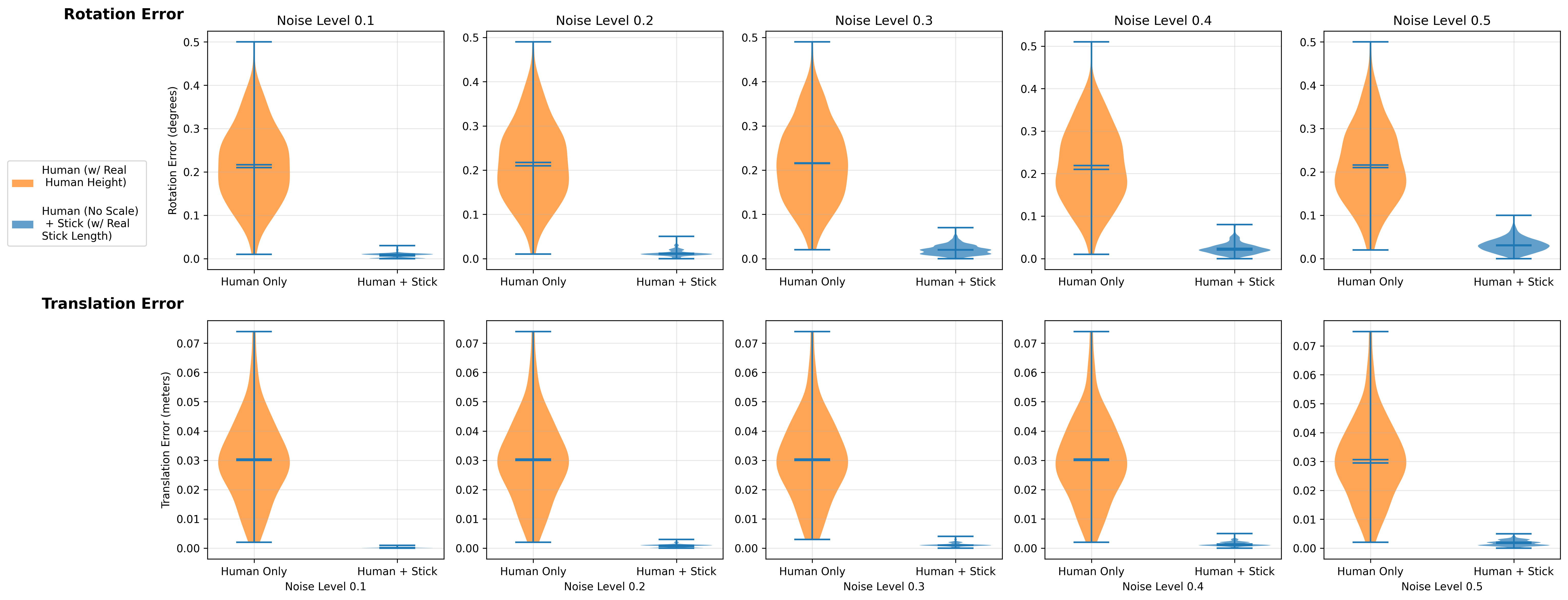}
  \vspace{-0.2cm}
  \caption{\textbf{Per-noise-level Error Distribution.} The human + stick approach yields lower errors and smaller variance in both rotation and translation.}
  \label{fig:noise_error_distributions}
\end{figure*}

\begin{table}[h!]
\centering
\caption{Ablation Study: Stage 3 Components.}
\vspace{-0.2cm}
\label{tab:stage3_ablation}
\setlength{\tabcolsep}{0.3pt}
\begin{tabular}{lcc}
\toprule
\textbf{Configuration} & \textbf{Avg Rot. Err. ($^\circ$)}$\downarrow$ & \textbf{Avg Trans. Err. (m)}$\downarrow$ \\
\midrule
\textbf{Ours (Full)} & \textbf{0.020} & \textbf{0.001} \\
\quad w/o Length Constraint & 0.091 &  0.011\\
\quad w/o Temporal Smoothness & \underline{0.022} & \underline{0.002} \\
\bottomrule
\end{tabular}
\vspace{-0.2cm}
\end{table}

Our approach resolves these issues by coupling the rigid, metric structure of the calibration stick with its motion-induced multi-view coverage. This dual constraint produces abundant cross-camera correspondences with consistent scale, which effectively stabilizes rotation estimation and eliminates metric drift in translation.

Moreover, the analysis of Fig.~\ref{fig:runtime_boxplot} highlights the scalability of the configuration process, showing how runtime increases with the number of cameras and the variability across trials. Our method's memory footprint is detailed in Tab.~\ref{tab:memory}. Initial Mem. reflects the static memory needed to load input data, while Peak Mem. indicates the maximum memory consumed during the final bundle adjustment stage. This peak is driven by the construction and factorization of the sparse Jacobian matrix. The results show that our method remains memory-efficient and scales well as the number of camera views increases.

\subsection{Ablation Studies}

We perform a detailed ablation study to isolate the contribution of each key component of our method. The results are summarized in Tabs.~\ref{tab:input_ablation} and~\ref{tab:stage3_ablation}, Figs.~\ref{fig:camera_error_distributions} and~\ref{fig:noise_error_distributions}.

As presented in Tab.~\ref{tab:input_ablation}, our analysis evaluates the performance of different input settings for multi-camera extrinsic calibration. A key challenge with using human subjects is the variability in height, which can be influenced by factors such as footwear and posture. This uncertainty is reflected in the results: a minor assumption error of $\pm 0.1$ m in human height increased the average rotation error by over $16\%$ and the average translation error by approximately $39\%$. While using a calibration stick with a known scale dramatically improves accuracy, it may not capture sufficient cross-camera geometric information, leading to suboptimal rotation estimation. Our proposed full method, which jointly utilizes the stick for its reliable scale and the human subject for rich motion cues, achieves the most robust performance. Notably, our approach reduces the average rotation error by a further $13.0\%$ compared to using the stick alone, while maintaining the same SOTA translation error of $0.001$ m. This demonstrates that integrating both information sources effectively mitigates their individual weaknesses, leading to superior calibration accuracy.

\begin{figure*}[th!]
  \centering
  \includegraphics[width=.95\linewidth]{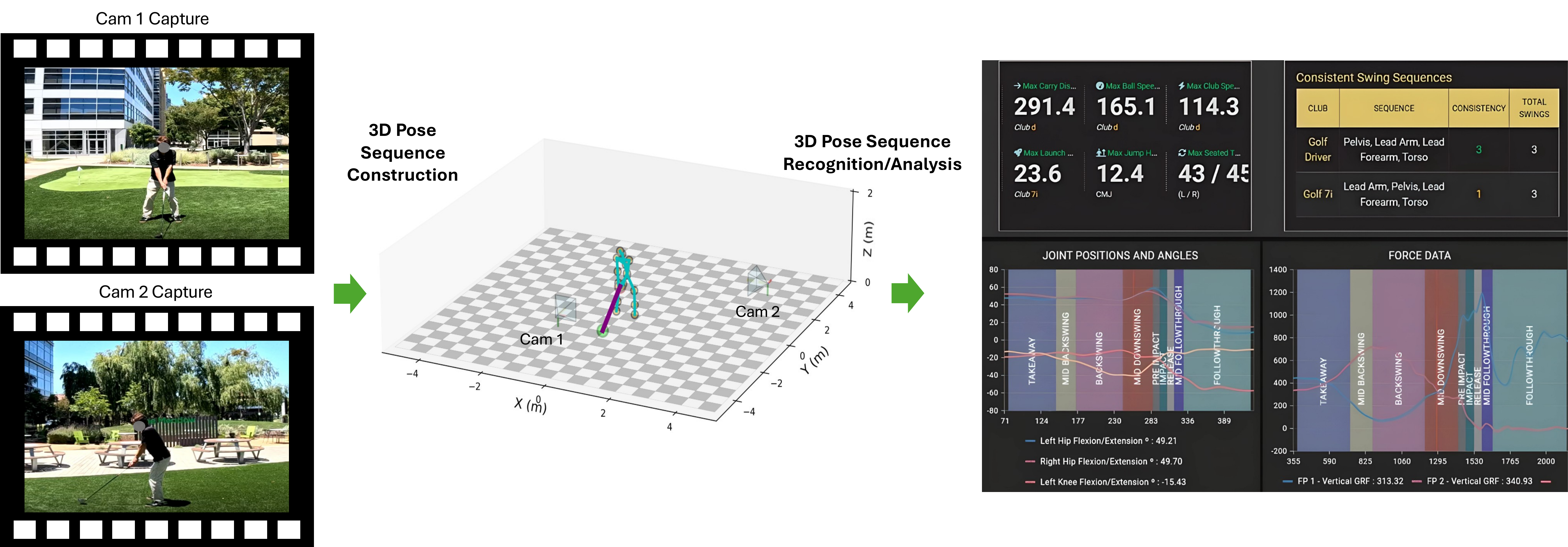}
  \vspace{-0.2cm}
  \caption{\textbf{Enabling precise 3D sports gesture analysis via tool-free self-calibration.} Our method utilizes rigid stick constraints to self-calibrate multi-camera setups (Left \& Center), enabling scale-aware 3D pose reconstruction. This supports robust downstream applications in sports analytics (Right), allowing for the accurate analysis of sport gestures and movement.}
  \label{fig:downstream_illustration}
\end{figure*}

Since our main contribution lies in introducing a given-scale stick to enable scale-aware multi-camera calibration, our approach differs from existing popular methods that rely solely on human poses~\cite{garau2020fast,lee2025spatiotemporal,lee2022extrinsic,patzold2022online,xu2021wide}. To highlight this distinction, we further report per-camera and per-noise-level errors to support a more detailed analysis. Specifically, we assessed the performance of our full `human + stick' method against the `Human Only' approach using 3 to 10 cameras (Fig.~\ref{fig:camera_error_distributions}). The results clearly show that our method consistently yields significantly lower errors and smaller variance in both rotation and translation, regardless of the number of cameras. The `Human Only' method, while showing some improvement with more cameras, consistently exhibits a wide error distribution, indicating instability. In contrast, our approach maintains high accuracy and precision, demonstrating its robustness to different multi-camera system configurations. We also tested the resilience of our method to varying levels of synthetic noise (estimated pose bias) in the input data, as shown in Fig.~\ref{fig:noise_error_distributions}. As noise levels increase, the `human + stick' approach exhibits only a slight broadening in both rotation and translation error distributions, yet retains a tightly clustered error profile with minimal degradation compared to the `human-only' approach, underscoring its robustness and reliability for real-world scenarios where noisy data are prevalent.

Finally, we evaluate the effectiveness of each component in Eq.~\eqref{eq:ba3_endpoints} for our Stage 3 joint optimization. The results are presented in Tab.~\ref{tab:stage3_ablation}. \textit{Effect of Stick Constraint:} Removing the stick length constraint from the final scale-aware bundle adjustment (w/o Length Constraint) significantly increases the camera pose error. This demonstrates that the constraint is essential for preserving the physical rigidity of the stick. \textit{Effect of Temporal Smoothness:} Removing the temporal smoothness prior (w/o Temporal Smoothness) leads to a noticeable increase in both rotation and translation errors. This indicates that the prior effectively regularizes the 3D trajectories against noise and contributes to more physically plausible motion.

\subsection{Real-scene Evaluation and Downstream Application}

To validate practical applicability beyond synthetic benchmarks, we evaluated our method in an uncontrolled outdoor scene using time-synchronized cameras (Fig.~\ref{fig:downstream_illustration}, Left). Real-world factors such as illumination variation, background clutter, and detector noise pose challenges that are largely absent in simulation. For 2D measurements, we fine-tuned a YOLOv11 model~\cite{yolo11_ultralytics} to detect human body keypoints and the two endpoints of the golf club, using only the athlete and the rigid implement as calibration references.
As shown in Fig.~\ref{fig:downstream_illustration} (Center), the proposed optimization recovers the relative camera extrinsics and reconstructs a coherent, metric-scale 3D human-club motion sequence from uncalibrated inputs. This 3D reconstruction serves as an interpretable intermediate representation that connects raw multi-view video to higher-level understanding.

The metric 3D pose sequence enables downstream sports analytics (Fig.~\ref{fig:downstream_illustration}, Right). We extract kinematic descriptors including joint angle trajectories, temporal phase segmentation of the swing, and sequence-level consistency metrics across repeated swings. The resulting curves and summaries are smooth and biomechanically plausible, providing strong evidence that the recovered extrinsics are sufficiently accurate for detailed motion analysis in real-world conditions.


\section{Conclusion}

We presented a novel and practical method for the tool-free extrinsic self-calibration of multi-camera systems in sports settings. Our key insight is to jointly leverage the rich, articulated motion of an athlete and the metric length constraint provided by a stick-like implement, such as a golf club or baseball bat. This combination effectively resolves the scale ambiguity inherent in calibration methods that rely on human motion alone. We introduced a robust three-stage optimization pipeline that first establishes a geometrically consistent reconstruction, then uses the stick's known length to recover the true metric scale, and finally refines all parameters in a scale-aware bundle adjustment with temporal priors. Our comprehensive experiments demonstrate that the proposed approach achieves SOTA performance, consistently yielding low rotation and translation errors. By eliminating the need for dedicated calibration hardware, our method provides a valuable and accessible tool for sports analytics, biomechanical studies, and performance tracking.


{\small
\bibliographystyle{IEEEtran}
\bibliography{egbib}

@inproceedings{xu2021wide,
  title={Wide-baseline multi-camera calibration using person re-identification},
  author={Xu, Yan and Li, Yu-Jhe and Weng, Xinshuo and Kitani, Kris},
  booktitle={Proceedings of the IEEE/CVF conference on computer vision and pattern recognition},
  pages={13134--13143},
  year={2021}
}

@article{lee2022extrinsic,
  title={Extrinsic camera calibration from a moving person},
  author={Lee, Sang-Eun and Shibata, Keisuke and Nonaka, Soma and Nobuhara, Shohei and Nishino, Ko},
  journal={IEEE Robotics and Automation Letters},
  volume={7},
  number={4},
  pages={10344--10351},
  year={2022},
  publisher={IEEE}
}

@article{jatesiktat2024multi,
  title={Multi-Camera Calibration Using Far-Range Dual-LED Wand and Near-Range Chessboard Fused in Bundle Adjustment},
  author={Jatesiktat, Prayook and Lim, Guan Ming and Ang, Wei Tech},
  journal={Sensors},
  volume={24},
  number={23},
  pages={7416},
  year={2024},
  publisher={MDPI}
}

@inproceedings{pan2024global,
  title={Global structure-from-motion revisited},
  author={Pan, Linfei and Bar{\'a}th, D{\'a}niel and Pollefeys, Marc and Sch{\"o}nberger, Johannes L},
  booktitle={European Conference on Computer Vision},
  pages={58--77},
  year={2024},
  organization={Springer}
}

@article{garau2020fast,
  title={Fast automatic camera network calibration through human mesh recovery},
  author={Garau, Nicola and De Natale, Francesco GB and Conci, Nicola},
  journal={Journal of Real-Time Image Processing},
  volume={17},
  number={6},
  pages={1757--1768},
  year={2020},
  publisher={Springer}
}

@inproceedings{yuhai2024enhanced,
  title={Enhanced Three-Axis Frame and Wand-Based Multi-Camera Calibration Method Using Adaptive Iteratively Reweighted Least Squares and Comprehensive Error Integration},
  author={Yuhai, Oleksandr and Cho, Yubin and Choi, Ahnryul and Mun, Joung Hwan},
  booktitle={Photonics},
  volume={11},
  number={9},
  pages={867},
  year={2024},
  organization={MDPI}
}

@article{marcon2017multicamera,
  title={Multicamera rig calibration by double-sided thick checkerboard},
  author={Marcon, Marco and Sarti, Augusto and Tubaro, Stefano},
  journal={IET Computer Vision},
  volume={11},
  number={6},
  pages={448--454},
  year={2017},
  publisher={Wiley Online Library}
}

@inproceedings{xing2017new,
  title={A new calibration technique for multi-camera systems of limited overlapping field-of-views},
  author={Xing, Ziran and Yu, Jingyi and Ma, Yi},
  booktitle={2017 IEEE/RSJ International Conference on Intelligent Robots and Systems (IROS)},
  pages={5892--5899},
  year={2017},
  organization={IEEE}
}

@inproceedings{atcheson2010caltag,
  title={Caltag: High precision fiducial markers for camera calibration.},
  author={Atcheson, Bradley and Heide, Felix and Heidrich, Wolfgang},
  booktitle={VMV},
  volume={10},
  pages={41--48},
  year={2010}
}

@article{pribanic2009comparison,
  title={A comparison between 2D plate calibration and wand calibration for 3D kinematic systems},
  author={Pribani{\'c}, Tomislav and Peharec, Stanislav and Medved, Vladimir},
  journal={Kinesiology},
  volume={41},
  number={2.},
  pages={147--155},
  year={2009},
  publisher={Kineziolo{\v{s}}ki fakultet}
}

@article{tabb2019multi,
  title={Multi-camera calibration with pattern rigs, including for non-overlapping cameras: CALICO},
  author={Tabb, Amy and Medeiros, Henry and Feldmann, Mitchell J and Santos, Thiago T},
  journal={arXiv preprint arXiv:1903.06811},
  year={2019}
}

@inproceedings{desai2018skeleton,
  title={Skeleton-based continuous extrinsic calibration of multiple RGB-D kinect cameras},
  author={Desai, Kevin and Prabhakaran, Balakrishnan and Raghuraman, Suraj},
  booktitle={Proceedings of the 9th ACM multimedia systems conference},
  pages={250--257},
  year={2018}
}

@article{lee2021robust,
  title={Robust extrinsic calibration of multiple RGB-D cameras with body tracking and feature matching},
  author={Lee, Sang-ha and Yoo, Jisang and Park, Minsik and Kim, Jinwoong and Kwon, Soonchul},
  journal={Sensors},
  volume={21},
  number={3},
  pages={1013},
  year={2021},
  publisher={MDPI}
}

@article{yang2024yowo,
  title={Yowo: You only walk once to jointly map an indoor scene and register ceiling-mounted cameras},
  author={Yang, Fan and Yamao, Sosuke and Kusajima, Ikuo and Moteki, Atsunori and Masui, Shoichi and Jiang, Shan},
  journal={IEEE Transactions on Circuits and Systems for Video Technology},
  year={2024},
  publisher={IEEE}
}

@inproceedings{theiner2023tvcalib,
  title={Tvcalib: Camera calibration for sports field registration in soccer},
  author={Theiner, Jonas and Ewerth, Ralph},
  booktitle={Proceedings of the IEEE/CVF winter conference on applications of computer vision},
  pages={1166--1175},
  year={2023}
}

@article{citraro2020real,
  title={Real-time camera pose estimation for sports fields},
  author={Citraro, Leonardo and M{\'a}rquez-Neila, Pablo and Savare, Stefano and Jayaram, Vivek and Dubout, Charles and Renaut, F{\'e}lix and Hasfura, Andres and Ben Shitrit, Horesh and Fua, Pascal},
  journal={Machine Vision and Applications},
  volume={31},
  number={3},
  pages={16},
  year={2020},
  publisher={Springer}
}

@article{tripicchio2022multi,
  title={Multi-camera extrinsic calibration for real-time tracking in large outdoor environments},
  author={Tripicchio, Paolo and D’Avella, Salvatore and Camacho-Gonzalez, Gerardo and Landolfi, Lorenzo and Baris, Gabriele and Avizzano, Carlo Alberto and Filippeschi, Alessandro},
  journal={Journal of Sensor and Actuator Networks},
  volume={11},
  number={3},
  pages={40},
  year={2022},
  publisher={MDPI}
}

@inproceedings{gossard2024ewand,
  title={eWand: An extrinsic calibration framework for wide baseline frame-based and event-based camera systems},
  author={Gossard, Thomas and Ziegler, Andreas and Kolmar, Levin and Tebbe, Jonas and Zell, Andreas},
  booktitle={2024 IEEE International Conference on Robotics and Automation (ICRA)},
  pages={14534--14540},
  year={2024},
  organization={IEEE}
}

@article{rameau2022mc,
  title={MC-Calib: A generic and robust calibration toolbox for multi-camera systems},
  author={Rameau, Francois and Park, Jinsun and Bailo, Oleksandr and Kweon, In So},
  journal={Computer Vision and Image Understanding},
  volume={217},
  pages={103353},
  year={2022},
  publisher={Elsevier}
}

@inproceedings{wang2025vggt,
  title={Vggt: Visual geometry grounded transformer},
  author={Wang, Jianyuan and Chen, Minghao and Karaev, Nikita and Vedaldi, Andrea and Rupprecht, Christian and Novotny, David},
  booktitle={Proceedings of the Computer Vision and Pattern Recognition Conference},
  pages={5294--5306},
  year={2025}
}

@article{keetha2025mapanything,
  title={MapAnything: Universal Feed-Forward Metric 3D Reconstruction},
  author={Keetha, Nikhil and M{\"u}ller, Norman and Sch{\"o}nberger, Johannes and Porzi, Lorenzo and Zhang, Yuchen and Fischer, Tobias and Knapitsch, Arno and Zauss, Duncan and Weber, Ethan and Antunes, Nelson and others},
  journal={arXiv preprint arXiv:2509.13414},
  year={2025}
}

@inproceedings{muller2025reconstructing,
  title={Reconstructing people, places, and cameras},
  author={M{\"u}ller, Lea and Choi, Hongsuk and Zhang, Anthony and Yi, Brent and Malik, Jitendra and Kanazawa, Angjoo},
  booktitle={Proceedings of the Computer Vision and Pattern Recognition Conference},
  pages={21948--21958},
  year={2025}
}

@inproceedings{patzold2022online,
  title={Online marker-free extrinsic camera calibration using person keypoint detections},
  author={P{\"a}tzold, Bastian and Bultmann, Simon and Behnke, Sven},
  booktitle={DAGM German Conference on Pattern Recognition},
  pages={300--316},
  year={2022},
  organization={Springer}
}

@article{lee2025spatiotemporal,
  title={Spatiotemporal Multi-Camera Calibration using Freely Moving People},
  author={Lee, Sang-Eun and Nishino, Ko and Nobuhara, Shohei},
  journal={IEEE Robotics and Automation Letters},
  year={2025},
  publisher={IEEE}
}

@inproceedings{usenko2018double,
  title={The double sphere camera model},
  author={Usenko, Vladyslav and Demmel, Nikolaus and Cremers, Daniel},
  booktitle={2018 International Conference on 3D Vision (3DV)},
  pages={552--560},
  year={2018},
  organization={IEEE}
}

@inproceedings{kanai2023robust,
  title={Robust self-supervised extrinsic self-calibration},
  author={Kanai, Takayuki and Vasiljevic, Igor and Guizilini, Vitor and Gaidon, Adrien and Ambrus, Rares},
  booktitle={2023 IEEE/RSJ International Conference on Intelligent Robots and Systems (IROS)},
  pages={1932--1939},
  year={2023},
  organization={IEEE}
}

@article{allegro2025calib3r,
  title={Calib3R: A 3D Foundation Model for Multi-Camera to Robot Calibration and 3D Metric-Scaled Scene Reconstruction},
  author={Allegro, Davide and Terreran, Matteo and Ghidoni, Stefano},
  journal={arXiv preprint arXiv:2509.08813},
  year={2025}
}

@article{umeyama2002least,
  title={Least-squares estimation of transformation parameters between two point patterns},
  author={Umeyama, Shinji},
  journal={IEEE Transactions on pattern analysis and machine intelligence},
  volume={13},
  number={4},
  pages={376--380},
  year={2002},
  publisher={IEEE}
}

@misc{mmpose2020,
    title={OpenMMLab Pose Estimation Toolbox and Benchmark},
    author={MMPose Contributors},
    howpublished = {\url{https://github.com/open-mmlab/mmpose}},
    year={2020}
}

@software{yolo11_ultralytics,
  author = {Glenn Jocher and Jing Qiu},
  title = {Ultralytics YOLOv11},
  version = {11.0.0},
  year = {2024},
  url = {https://github.com/ultralytics/ultralytics},
  orcid = {0000-0001-5950-6979, 0000-0003-3783-7069},
  license = {AGPL-3.0}
}

@article{kim2018charuco,
  title={{Charuco Board-Based Omnidirectional Camera Calibration Method}},
  author={Kim, Dae-Yong and Kim, Ji-Hyeok and Kim, Kyoung-Tae},
  journal={Sensors},
  volume={18},
  number={12},
  pages={421},
  year={2018},
  publisher={MDPI}
}

@inproceedings{bolsee2020device,
  title={A Device for Capturing Inward-Looking Spherical Light Fields},
  author={Bols{\'e}e, Quentin and Darwish, Walid and Bonatto, Daniele and Lafruit, Gauthier and Munteanu, Adrian},
  booktitle={2020 International Conference on 3D Immersion (IC3D)},
  pages={1--5},
  year={2020},
  organization={IEEE}
}

@inproceedings{kedilioglu2025pricosa,
  title={PrIcosa: High-Precision 3D Camera Calibration with Non-Overlapping Field of Views},
  author={Kedilioglu, Oguz and Nova, Tasnim Tabassum and Landesberger, Martin and Wang, Lijiu and Hofmann, Michael and Franke, J{\"o}rg and Reitelsh{\"o}fer, Sebastian},
  booktitle={Proceedings of the 20th International Joint Conference on Computer Vision, Imaging and Computer Graphics Theory and Applications-(Volume 2)},
  pages={801--809},
  year={2025},
  organization={SciTePress}
}

@inproceedings{strauss2014calibrating,
  title={Calibrating multiple cameras with non-overlapping views using coded checkerboard targets},
  author={Strau{\ss}, Tobias and Ziegler, Julius and Beck, Johannes},
  booktitle={17th international IEEE conference on intelligent transportation systems (ITSC)},
  pages={2623--2628},
  year={2014},
  organization={IEEE}
}

@article{prible2024caliscope,
  title={Caliscope: GUI Based Multicamera Calibration and Motion Tracking},
  author={Prible, Donald},
  journal={Journal of Open Source Software},
  volume={9},
  number={102},
  pages={7155},
  year={2024}
}

@inproceedings{chen2019sports,
  title={Sports camera calibration via synthetic data},
  author={Chen, Jianhui and Little, James J},
  booktitle={Proceedings of the IEEE/CVF conference on computer vision and pattern recognition workshops},
  pages={0--0},
  year={2019}
}

@inproceedings{chauve2010robust,
  title={Robust piecewise-planar 3D reconstruction and completion from large-scale unstructured point data},
  author={Chauve, Anne-Laure and Labatut, Patrick and Pons, Jean-Philippe},
  booktitle={2010 IEEE computer society conference on computer vision and pattern recognition},
  pages={1261--1268},
  year={2010},
  organization={IEEE}
}

@phdthesis{wu2008multi,
  title={Multi-view hockey tracking with trajectory smoothing and camera selection},
  author={Wu, Lan},
  year={2008},
  school={University of British Columbia}
}

@article{ren2010multi,
  title={Multi-camera video surveillance for real-time analysis and reconstruction of soccer games},
  author={Ren, Jinchang and Xu, Ming and Orwell, James and Jones, Graeme A},
  journal={Machine Vision and Applications},
  volume={21},
  number={6},
  pages={855--863},
  year={2010},
  publisher={Springer}
}

@article{luo2013feature,
  title={Feature extraction and representation for distributed multi-view human action recognition},
  author={Luo, Jiajia and Wang, Wei and Qi, Hairong},
  journal={IEEE Journal on Emerging and Selected Topics in Circuits and Systems},
  volume={3},
  number={2},
  pages={145--154},
  year={2013},
  publisher={IEEE}
}

@article{zhang2020multi,
  title={Multi-camera multi-player tracking with deep player identification in sports video},
  author={Zhang, Ruiheng and Wu, Lingxiang and Yang, Yukun and Wu, Wanneng and Chen, Yueqiang and Xu, Min},
  journal={Pattern Recognition},
  volume={102},
  pages={107260},
  year={2020},
  publisher={Elsevier}
}

@article{wu2022survey,
  title={A survey on video action recognition in sports: Datasets, methods and applications},
  author={Wu, Fei and Wang, Qingzhong and Bian, Jiang and Ding, Ning and Lu, Feixiang and Cheng, Jun and Dou, Dejing and Xiong, Haoyi},
  journal={IEEE Transactions on Multimedia},
  volume={25},
  pages={7943--7966},
  year={2022},
  publisher={IEEE}
}

@article{wang2024diffusion,
  title={Diffusion Convolution Neural Network-Based Multiview Gesture Recognition for Athletes in Dynamic Scenes},
  author={Wang, Qingyun and Li, Hua},
  journal={Journal of Circuits, Systems and Computers},
  volume={33},
  number={06},
  pages={2450114},
  year={2024},
  publisher={World Scientific}
}

@article{yang2024enhancing,
  title={Enhancing multi-camera gymnast tracking through domain knowledge integration},
  author={Yang, Fan and Odashima, Shigeyuki and Masui, Shoichi and Kusajima, Ikuo and Yamao, Sosuke and Jiang, Shan},
  journal={IEEE Transactions on Circuits and Systems for Video Technology},
  year={2024},
  publisher={IEEE}
}

@article{noorbhai2025conceptual,
  title={A conceptual framework and review of multi-method approaches for 3D markerless motion capture in sports and exercise},
  author={Noorbhai, Habib and Moon, Sanghee and Fukushima, Takashi},
  journal={Journal of sports sciences},
  volume={43},
  number={12},
  pages={1167--1174},
  year={2025},
  publisher={Taylor \& Francis}
}

@phdthesis{taylor2025biomechanical,
  title={Biomechanical Golf Swing Analysis using Markerless Three-Dimensional Skeletal Tracking through Truncation-Robust Heatmaps},
  author={Taylor, Benjamin F and others},
  year={2025},
  school={Massachusetts Institute of Technology}
}

@article{javerliat2025kineo,
  title={Kineo: Calibration-Free Metric Motion Capture From Sparse RGB Cameras},
  author={Javerliat, Charles and Raimbaud, Pierre and Lavou{\'e}, Guillaume},
  journal={arXiv preprint arXiv:2510.24464},
  year={2025}
}
}

\end{document}